\pdfoutput=1

\documentclass[11pt,a4paper]{article}
\usepackage[hyperref]{eacl2021}
\usepackage{times}
\usepackage{latexsym}

\usepackage{microtype}

\aclfinalcopy 

\usepackage{pifont}
\usepackage{tabularx}
\usepackage{booktabs}
\usepackage{graphicx}
\usepackage{xcolor}

\title{TMR: Evaluating NER Recall on Tough Mentions}

\author{Jingxuan Tu \\
    Brandeis University \\
    \texttt{jxtu@brandeis.edu} \\\And
    Constantine Lignos \\
    Brandeis University \\
    \texttt{lignos@brandeis.edu}
}

\date{}

\begin{document}
\maketitle

\begin{abstract}
We propose the \emph{Tough Mentions Recall} (TMR) metrics to supplement traditional named entity recognition (NER) evaluation by examining recall on specific subsets of ``tough'' mentions: \emph{unseen mentions}, those whose tokens or token/type combination were not observed in training, and \emph{type-confusable mentions}, token sequences with multiple entity types in the test data.
We demonstrate the usefulness of these metrics by evaluating corpora of English, Spanish, and Dutch using five recent neural architectures.
We identify subtle differences between the performance of BERT and Flair on two English NER corpora and identify a weak spot in the performance of current models in Spanish.
We conclude that the TMR metrics enable differentiation between otherwise similar-scoring systems and identification of patterns in performance that would go unnoticed from overall precision, recall, and F1.
\end{abstract}

\section{Introduction}

    For decades, the standard measures of performance for named entity recognition (NER) systems have been precision, recall, and F1 computed over entity mentions.\footnote{
      We use the term \emph{mention} to refer to a specific annotated reference to a named entity---a  span of tokens (\emph{token sequence}) and an entity type.
      We reserve the term \emph{entity} for the referent, e.g. the person being named.
      The traditional NER F1 measure is computed over mentions (``phrase'' F1).
    }
    NER systems are primarily evaluated using exact match\footnote{
        While partial match metrics have been used \citep[e.g.][]{chinchor-sundheim-1993-muc, chinchor-1998-appendix, doddington-etal-2004-automatic, segura-bedmar-etal-2013-semeval}, exact matching is still most commonly used, and the only approach we explore.}
    F1 score, micro-averaged across mentions of all entity types.
    While per-entity-type scores available from the \texttt{conlleval} scorer \citep{tjong-kim-sang-2002-introduction} are often reported, there are no widely-used diagnostic metrics that further analyze the performance of NER systems and allow for separation of systems close in F1.
    
    This work proposes \emph{Tough Mentions Recall} (TMR), a set of metrics that provide a fine-grained analysis of the mentions that are likely to be most challenging for a system: \emph{unseen mentions}, ones that are present in the test data but not the training data, and \emph{type-confusable mentions}, ones that appear with multiple types in the test set.
    We evaluate the performance of five recent popular neural systems on English, Spanish and Dutch data using these fine-grained metrics.
    We demonstrate that TMR metrics enable differentiation between otherwise similar-scoring systems, and the model that performs best overall might not be the best on the tough mentions.
    Our NER evaluation tool is publicly available via a GitHub
    repository.\footnote{\url{https://github.com/jxtu/EvalNER}}
    
\section{Related Work}
    
    Previous work in NER and sequence labeling has examined performance on out-of-vocabulary (OOV) tokens and rare or unseen entities.
    \citet{ma-hovy-2016-end} and \citet{yang-etal-2018-design} evaluate system performance on mentions containing tokens not present in the pretrained embeddings or training data.
    Such analysis can be used broadly---\citeauthor{ma-hovy-2016-end} perform similar analyses for part of speech tagging and NER---and can guide system design around the handling of those tokens.

    \citet{augenstein17} present a thorough analysis of the generalization abilities of NER systems, quantifying the performance gap between seen and unseen mentions, among many other factors.
    Their work predates current neural NER models; the newest model they use in their evaluation is SENNA \citep{collobert2011natural}. 
    While prior work has considered evaluation on unseen mentions, it has focused on experimenting on English data, and the definition of ``unseen'' has focused on the tokens themselves being unseen (\textsc{Unseen-Tokens} in our work).
    We use the umbrella of ``tough mentions'' to cover a number of possible distinctions that can be made with regards to how unseen test set data is, and we experiment on multiple languages.

    \citet{tse-ner} propose an iterative approach for long-tail entity extraction, focusing on entities of two specific types in the scientific domain.
    \citet{derczynski-etal-2017-results} propose evaluation on a set of unique mentions, which emphasizes the ability of a system to recognize rarer entities.
    As entities and their types change quickly \citep{DERCZYNSKI201532}, recall on emerging entities is becoming a more critical measure in evaluating progress. 
    \citet{ribeiro-etal-2020-beyond} propose \textsc{CheckList}, which can be applied to NER by using invariance tests; for example, replacing a mention with another one of the same entity type should not affect the output of the model.
    \citet{Fu_Liu_Zhang_2020} evaluate the generalization of NER models through breakdown tests, annotation errors and dataset bias.
    They examine the performance on subsets of entities based on the entity coverage rate between train and test set. They also release \textit{ReCoNLL}, a revised version of CoNLL-2003 English with fewer annotation errors which we use in this work.

\section{Unseen and Type-confusable Mentions}

    \begin{table}[tb]
        \centering
        \resizebox{\linewidth}{!}{
        \begin{tabular}{lccc}
            \toprule
            \textsc{Training Set} & \multicolumn{3}{l}{\textbf{Newcastle\textsubscript{[LOC]}} is a city in the \textbf{UK\textsubscript{[LOC]}}.} \\
            \midrule
            \textsc{Test Set} & \multicolumn{3}{l}{\textbf{John Brown\textsubscript{[PER]}}, the \textbf{Newcastle\textsubscript{[ORG]}}} \\
            {} &   \multicolumn{3}{l}{star from the \textbf{UK\textsubscript{[LOC]}}, has\ldots} \\
            \bottomrule
            \\
            \toprule
            {} & \textbf{Newcastle} & \textbf{John} & \textbf{UK\textsubscript{[LOC]}} \\
            {} & \textbf{\textsubscript{[ORG]}}  & \textbf{Brown\textsubscript{[PER]}} & {} \\
            \midrule
            \textsc{Seen}        &  {} & {} & \ding{52}\\
            \textsc{Unseen-Type}        &  \ding{52} & {} & {} \\
            \textsc{Unseen-Tokens}       &  {} & \ding{52} & {} \\
            \textsc{Unseen-Any}        &   \ding{52} & \ding{52} & {} \\
            \bottomrule
        \end{tabular}
        }
        \caption{Example data and how mentions would be classified into unseen and type-confusable mention sets}
        \label{OOV_examples}
    \end{table}

    \subsection{Unseen Mentions}

    Given annotated NER data divided into a fixed train/development/test split, we are interested in the relationship between the mentions of the training and test sets.
    We classify mentions into three mutually-exclusive sets described in Table~\ref{OOV_examples}: \textsc{Seen}, \textsc{Unseen-Type}, and \textsc{Unseen-Tokens}, and a superset \textsc{Unseen-Any} that is the union of \textsc{Unseen-Type} and \textsc{Unseen-Tokens}.
    \textbf{UK\textsubscript{[LOC]}} appears in both the training and test set, so it is a \textsc{Seen} mention.
    As there is no mention consisting of the token sequence \emph{John Brown} annotated as any type in the test set, \textbf{John Brown\textsubscript{[PER]}} is an \textsc{Unseen-Tokens} mention.\footnote{The matching criterion for the token sequence is case sensitive, requires an exact---not partial---match, and only considers mentions. \textbf{John Henry Brown\textsubscript{[PER]}}, \textbf{john brown\textsubscript{[PER]}}, or unannotated \emph{John Brown} appearing in the training set would not make \textbf{John Brown\textsubscript{[PER]}} a seen mention.}
    While there is no mention with the tokens and type \textbf{Newcastle\textsubscript{[ORG]}} in the training data, the token sequence \emph{Newcastle} appears as a mention, albeit with a different type (LOC).
    \textbf{Newcastle\textsubscript{[ORG]}} is an \textsc{Unseen-Type} mention as the same token sequence has appeared as a mention, but not with the type ORG.

    \subsection{Type-confusable Mentions}
    Token sequences that appear as mentions with multiple types in the test set form another natural set of challenging mentions.
    If \textbf{Boston\textsubscript{[LOC]}}, the city, and \textbf{Boston\textsubscript{[ORG]}}, referring to a sports team\footnote{For example: \emph{\textbf{Boston\textsubscript{[ORG]}} won the World Series in 2018}.} are both in the test set, we consider all mentions of exactly the token sequence \emph{Boston} to be \emph{type-confusable mentions} (TCMs), members of \textsc{TCM-All}.
    We can further divide this set based on whether each mention is unseen. \textsc{TCM-Unseen} is the intersection of \textsc{TCM-All} and \textsc{Unseen-Token}; \textsc{TCM-Seen} is the rest of \textsc{TCM-All}.

    Unlike \citet{Fu_Liu_Zhang_2020}, who explore token sequences that occur with different types in the training data, we base our criteria for TCMs around type variation in the test data.
    Doing so places the focus on whether the model can correctly produce multiple types in the output, as opposed to how it reacted to multiple types in the input.
    Also, if type confusability were based on the training data, it would be impossible to have \textsc{TCM-Unseen} mentions, as the fact that they are type confusable in the training data means they have been seen at least twice in training and thus cannot be considered unseen.
    As our metrics compute subsets over the gold standard entities, it is natural to only measure recall and not precision on those subsets, as it is not clear exactly which false positives should be considered in computing precision.

    \subsection{Data Composition}
    \label{sec:data}

    \begin{table}[tb]
        \centering
                \resizebox{\linewidth}{!}{
            \begin{tabular}{lrrrrr}
                \toprule
                Set & LOC & ORG & PER & MISC & ALL \\
                \midrule
                \textsc{Unseen-Any}     & 17.9 & 45.9 & \textbf{85.3} & 35.5 & 47.6 \\
                \textsc{Unseen-Tok.}    & 17.5 & 41.6 & \textbf{85.1} & 35.1 & 46.1 \\
                \textsc{Unseen-Type}    & 0.4 & \textbf{4.3} & 0.2 & 0.4 & 1.5 \\
                \midrule
                \textsc{TCM-All}        & 7.1 & \textbf{13.7} & 0.4 & 1.0 & 6.3 \\
                \textsc{TCM-Seen}       & 5.4 & \textbf{9.5} & 0.4 & 1.0 & 4.6 \\
                \textsc{TCM-Unseen}     & 1.7 & \textbf{4.2} & 0.0 & 0.0 & 1.7 \\
                \midrule
                All (Count)             & 1,668 & 1,661 & 1,617 & 702 & 5,648 \\
                \bottomrule
            \end{tabular}}
        \caption{Percentage of all mentions in each subset, with total mentions in the final row (ReCoNLL English)}
        \label{mention_percentage_re}
    \end{table}
    
    \begin{table}[tb]
        \centering
        \resizebox{\linewidth}{!}{
            \begin{tabular}{lrrrrr}
                \toprule
                Set & LOC & ORG & PER & MISC & ALL \\
                \midrule
                \textsc{Unseen-Any}        & 24.4 & 30.8 & \textbf{68.9} & 60.9 & 39.6 \\
                \textsc{Unseen-Tok.} & 22.4 & 29.2 & \textbf{67.1} & 58.8 & 37.8 \\
                \textsc{Unseen-Type}        & 2.0 & 1.6 & 1.8 & \textbf{2.1} & 1.8 \\
                \midrule
                \textsc{TCM-All}        & \textbf{23.3} & 7.5 & 1.1 & 4.7 & 10.7 \\
                \textsc{TCM-Seen}   & \textbf{22.6} & 6.8 & 0.8 & 4.1 & 10.1 \\
                \textsc{TCM-Unseen} & \textbf{0.7} & \textbf{0.7} & 0.3 & 0.6 & 0.6 \\
                \midrule
                All (Count)       & 1,084 & 1,400 & 735 & 340 & 3,559 \\

                \bottomrule
            \end{tabular}}
        \caption{Percentage of all mentions in each subset, with total mentions in the final row (CoNLL-2002 Spanish)}
        \label{spanish_mention_percentage}
    \end{table}

        \begin{table}[tb]
        \centering
        \resizebox{\linewidth}{!}{
            \begin{tabular}{lrrrrr}
                \toprule
                Set & LOC & ORG & PER & MISC & ALL \\
                \midrule
                \textsc{Unseen-Any}        & 36.8 & 52.2 & \textbf{72.6} & 51.2 & 54.6 \\
                \textsc{Unseen-Tok.} & 36.8 & 52.1 & \textbf{72.5} & 50.9 & 54.4 \\
                \textsc{Unseen-Type}        & 0.0 & 0.1 & 0.1 & \textbf{0.3} & 0.2 \\
                \midrule
                \textsc{TCM-All}        & 0.1 & 0.0 & 0.2 & \textbf{0.3} & 0.2 \\
                \textsc{TCM-Seen}   & \textbf{0.1} & 0.0 & \textbf{0.1} & 0.0 & 0.1 \\
                \textsc{TCM-Unseen} & 0.0 & 0.0 & 0.1 & \textbf{0.3} & 0.1 \\
                \midrule
                All (Count)        & 774 & 882 & 1,098 & 1,187 & 3,941 \\
                \bottomrule
            \end{tabular}}
        \caption{Percentage of all mentions in each subset, with total mentions in the final row (CoNLL-2002 Dutch)}
        \label{dutch_mention_percentage}
    \end{table}

    We evaluate using the ReCoNLL English \citep{Fu_Liu_Zhang_2020}, OntoNotes 5.0 English \citep[using data splits from \citealt{pradhan-etal-2013-towards}]{OntoNotes}, CoNLL-2002 Dutch, and CoNLL-2002 Spanish \citep{tjong-kim-sang-2002-introduction} datasets.
    We use ReCoNLL \citep{Fu_Liu_Zhang_2020} in our analysis instead of the CoNLL-2003 English data \citep{tjong-kim-sang-de-meulder-2003-introduction} to improve accuracy as it contains a number of corrections. 

    Tables~\ref{mention_percentage_re}, \ref{spanish_mention_percentage}, and \ref{dutch_mention_percentage} give the total mentions of each entity type and the percentage that fall under the proposed unseen and TCM subsets for the three CoNLL datasets.\footnote{Tables for OntoNotes 5.0 English are provided in the appendix (Tables~\ref{onto-percentage1}-\ref{onto-percentage2}).}
    Across the three languages, 39.6\%--54.6\% of mentions are unseen, with the highest rate coming from PER mentions.
    \textsc{Unseen-Type} contains under 2\% of mentions in English and Spanish and almost no mentions in Dutch; it is rare for a token sequence to only appear in training with types that do not appear with it in the test data.
    
    Similarly, TCMs appear in the English (10.7\%) and Spanish (6.3\%) data, but almost never in Dutch (0.2\%).
    The differences across languages with regards to TCMs may reflect morphology or other patterns that prevent the same token sequence from appearing with multiple types, but they could also be caused by the topics included in the data.
    In English, the primary source of TCMs is the use of city names as sports organizations, creating LOC-ORG confusion.

\section{Results}
\label{exp-results}

    \subsection{Models and Evaluation}

    We tested five recent mainstream NER neural architectures that either
    achieved the state-of-the-art performance previously or are widely
    used among the research community.\footnote{We could not include a recent system by \citet{baevski-etal-2019-cloze} because it was not made publicly available.}
    The models are
    \textsc{CharCNN+WordLSTM+CRF}\footnotemark (\textsc{CharCNN}),
    \textsc{CharLSTM+WordLSTM+CRF}\footnotemark[\value{footnote}] (\textsc{CharLSTM}),
    \footnotetext{Using the NCRF++ \citep{yang-zhang-2018-ncrf} implementations:
        \url{https://github.com/jiesutd/NCRFpp}.}
    \textsc{Cased BERT-Base}\footnote{NER implementation from \url{https://github.com/kamalkraj/BERT-NER}.} \citep{devlin-etal-2019-bert}, \textsc{BERT-CRF}\footnote{A Cased BERT-Base Model with an additional CRF layer.} \citep{bertcrf}, and \textsc{Flair} \citep{akbik-etal-2018-contextual}.\footnote{\url{https://github.com/flairNLP/flair}}
    
    We trained all the models using the training set of each dataset.
    We fine-tuned English Cased BERT-Base, Dutch \citep{bert-dutch} and Spanish \citep{bert-spanish} BERT models and used the model from epoch 4 after comparing development set performance for epochs 3, 4, and 5. We also fine-tuned BERT-CRF models using the training data, and used the model from the epoch where development set performance was the best within the maximum of 16 epochs.   

    All models were trained five times each on a single NVIDIA TITAN RTX GPU. The mean and standard deviation of scores over five training runs are reported for each model.
    It took approximately 2 hours to train each of \textsc{Flair} and NCRF++ on each of the CoNLL-2002/3 datasets, 12 hours to train \textsc{Flair}, and 4 hours to train NCRF++ on OntoNotes 5.0 English.
    It took less than an hour to fine-tune BERT or BERT-CRF models on each dataset.
    Hyperparameters for Spanish and Dutch models implemented using NCRF++ were taken from \citet{lample-etal-2016-neural}. \textsc{Flair} does not provide hyperparameters for training CoNLL-02 Spanish, so we used those for CoNLL-02 Dutch.
    We did not perform any other hyperparameter tuning.

    
      \begin{table}[tb]
        \small
        \centering
        \resizebox{\linewidth}{!}{
            \begin{tabular}{lrrr}
                \toprule
                Model & Precision & Recall & F1 \\
                \midrule
                \textsc{CharLSTM} & 91.92 ($\pm$0.29) & 91.90 ($\pm$0.31) & 91.91 ($\pm$0.28) \\
                \textsc{CharCNN}  & 92.13 ($\pm$0.18) & 91.93 ($\pm$0.18) & 92.03 ($\pm$0.17) \\
                \textsc{Flair}    & \textbf{93.00 ($\pm$0.15)} & \textbf{93.66 ($\pm$0.08)} &  \textbf{93.33 ($\pm$0.12)} \\
                 \textsc{BERT} & 91.04 ($\pm$0.11) & 92.36 ($\pm$0.13) & 91.70 ($\pm$0.14) \\
                 \textsc{BERT-CRF} & 91.13 ($\pm$0.15) & 92.29 ($\pm$0.04) & 91.70 ($\pm$0.08) \\
                \bottomrule
            \end{tabular}}
        \caption{Standard P/R/F1 (ReCoNLL-2003 English)}
        \label{standard_all_type_reconll}
    \end{table}

    \begin{table}[tb]
        \small
        \centering
        \resizebox{\linewidth}{!}{
            \begin{tabular}{lrrr}
                \toprule
                Model & Precision & Recall & F1 \\
                \midrule
                \textsc{CharLSTM} & 87.12 ($\pm$0.42) & 86.38 ($\pm$0.36) & 86.90 ($\pm$0.40)\\
                \textsc{CharCNN}  & 86.94 ($\pm$0.27) & 86.28 ($\pm$0.33) & 86.61 ($\pm$0.25) \\
                \textsc{Flair}    & \textbf{88.56 ($\pm$0.12)} & 89.42 ($\pm$0.09)  &  \textbf{88.99 ($\pm$0.10)} \\
                \textsc{BERT}     & 87.52 ($\pm$0.09) & \textbf{89.84 ($\pm$0.12)} & 88.67 ($\pm$0.10) \\
                \textsc{BERT-CRF}     & 87.29 ($\pm$0.33) & 89.32 ($\pm$0.19) & 88.29 ($\pm$0.26) \\

                \bottomrule
            \end{tabular}}
        \caption{Standard P/R/F1 (OntoNotes 5.0 English)}
        \label{standard_all_type_onto}
    \end{table}

    \subsection{Baseline Results}
    We first examine the performance of these systems under standard evaluation measures.
    Tables~\ref{standard_all_type_reconll} and \ref{standard_all_type_onto} give performance on ReCoNLL and OntoNotes 5.0 English datasets using standard P/R/F1.
    In English, Flair attains the best F1 in both datasets, although BERT attains higher recall for OntoNotes.\footnote{
    We are not aware of any open-source implementation capable of matching the F1 of 92.4 reported by \citet{devlin-etal-2019-bert}. The gap between published and reproduced performance likely stems from the usage of the ``maximal document context,'' while reimplementations process sentences independently, as is typical in NER.
    Performance of Flair is slightly worse than that reported in the original paper because we did not use the development set as additional training data.}
    
    BERT attains the highest F1 in Dutch (91.26) and Spanish (87.36); due to space limitations, tables are provided in the appendix (Tables~\ref{standard_all_type_dutch}-\ref{standard_all_type_spanish}).
    BERT-CRF performs similar or slightly worse than BERT in all languages, but generally attains lower standard deviation in multiple training runs, which suggests greater stability from using a CRF for structured predictions.
    The same observation also holds for Flair which also uses a CRF layer.
    We are not aware of prior work showing results from using BERT-CRF on English, Spanish, and Dutch.
    \citet{bertcrf} shows that the combination of Portuguese BERT Base and CRF does not show better performance than bare BERT Base, which agrees with our observations.
    F1 rankings are otherwise similar across languages. The performance of CharLSTM and CharCNN cannot be differentiated in English, but CharLSTM substantially outperforms CharCNN in Spanish (+2.53) and Dutch (+2.15).

    \subsection{TMR for English}

    We explore English first and in greatest depth because its test sets are much larger than those of the other languages we evaluate, and we have multiple well-studied test sets for it.
    Additionally, the CoNLL-2003 English test data is from a later time than the training set, reducing train/test similarity.

        \begin{table}[tb]
        \centering
        \small
        \resizebox{\linewidth}{!}{
            \begin{tabular}{lrrrr}
                \toprule
                {} & \textsc{All} & \textsc{TCM-} & \textsc{TCM-} & \textsc{TCM-} \\
                Model &  & \textsc{All} & \textsc{Seen} & \textsc{Unseen} \\
                \midrule
                \textsc{CharLSTM}  & 91.90 & 85.52 ($\pm$1.09) & 87.36 ($\pm$0.70) & 80.61 ($\pm$3.00)\\
                \textsc{CharCNN} & 91.93 & 85.58 ($\pm$1.08) & 87.55 ($\pm$1.11) & 80.36 ($\pm$3.37)\\
                \textsc{Flair}   & \textbf{93.66} & \textbf{88.47 ($\pm$0.51)} & \textbf{89.75 ($\pm$0.73)} & \textbf{87.76 ($\pm$1.86)} \\
                \textsc{BERT}    & 92.36 & 88.28 ($\pm$0.74) & 89.69 ($\pm$0.89) & 85.46 ($\pm$1.74) \\
                \textsc{BERT-CRF}   & 92.29 & 87.02 ($\pm$0.71) & 89.43 ($\pm$0.76) & 79.59 ($\pm$1.25) \\
                \bottomrule
            \end{tabular}}
        \caption{Recall over all mentions and each type-confusable mention subset (ReCoNLL-2003 English)}
        \label{tce_all_type_reconll}
    \end{table}
    
    \paragraph{Revised CoNLL English.}

    One of the advantages of evaluating using TMR metrics is that systems can be differentiated more easily.
    Table~\ref{tce_all_type_reconll} gives recall for type-confusable mentions (TCMs) on ReCoNLL English.
    As expected, recall for TCMs is lower than overall recall, but more importantly, recall is less tightly-grouped over the TCM subsets (range of 8.17) than all mentions (1.76).
    This spread allows for better differentiation, even though there is a higher standard deviation for each score.
    For example, BERT-CRF generally performs very similarly to BERT, but scores 5.87 points lower for \textsc{TCM-Unseen}, possibly due to how the CRF handles lower-confidence predictions differently \citep{lignos-kamyab-2020-build}.
    Flair has the highest all-mentions recall and the highest recall for TCMs, suggesting that when type-confusable mentions have been seen in the training data, it is able to effectively disambiguate types based on context.
    
    \begin{table}[t]
        \centering
        \small
        \resizebox{\linewidth}{!}{
        \begin{tabular}{lrrrr}
        \toprule
        Model & \textsc{All} & \textsc{U-Any} & \textsc{U-Tok.} & \textsc{U-Type} \\
        \midrule
        \textsc{CharLSTM}  & 91.90 & 86.94 ($\pm$0.58) & 87.32 ($\pm$0.63) & 75.29 ($\pm$2.54) \\
        \textsc{CharCNN}   & 91.93 & 87.06 ($\pm$0.21) & 87.48 ($\pm$0.18) & 74.41 ($\pm$1.48) \\
        \textsc{Flair}     & \textbf{93.66} & \textbf{89.93 ($\pm$0.25)} & \textbf{90.31 ($\pm$0.19)} & 78.53 ($\pm$2.94) \\
        \textsc{BERT}      & 92.36  & 87.94 ($\pm$0.29) & 88.02 ($\pm$0.31) & \textbf{85.29 ($\pm$2.04)} \\
        \textsc{BERT-CRF}      & 92.29  & 87.55 ($\pm$0.14) & 87.73 ($\pm$0.12) & 82.12 ($\pm$1.53) \\

        \bottomrule
        \end{tabular}}
        \caption{Recall over all mentions and each unseen (\textsc{U-}) mention subset (ReCoNLL-2003 English)}
        \label{oov_all_type_reconll}
    \end{table}

    Table~\ref{oov_all_type_reconll} gives recall for unseen mentions.
    Although Flair attains higher overall recall, BERT attains higher recall on \textsc{Unseen-Type}, the set on which all models perform their worst.
    While there are few (85) mentions in this set, making assessment of statistical reliability challenging, this set allows us to identify an advantage for BERT in this specific subset: a BERT-based NER model is better able to produce a novel type for a token sequence only seen with other types in the training data.

    \begin{table}[t]
        \small
        \centering
        \resizebox{\linewidth}{!}{
        \begin{tabular}{lrrr}
            \toprule
            Model & \textsc{All} & \textsc{TCM-All} & \textsc{TCM-Seen} \\ 
            \midrule
            \textsc{CharLSTM} & 86.38 & 80.65 ($\pm$0.46) & 82.24 ($\pm$0.46) \\ 
            \textsc{CharCNN} & 86.28 & 79.80 ($\pm$0.41) & 81.49 ($\pm$0.40) \\ 
            \textsc{Flair} & 89.42 & \textbf{86.00 ($\pm$0.44)} & \textbf{87.39 ($\pm$0.51)} \\ 
            \textsc{BERT}  & \textbf{89.84} & 84.72 ($\pm$0.18) & 85.61 ($\pm$0.00) \\ 
             \textsc{BERT-CRF}    & 89.32 & 85.46 ($\pm$0.40) & 86.83 ($\pm$0.46) \\ 
            \bottomrule
        \end{tabular}}
        \caption{Recall over all mentions and each type-confusable mention subset (OntoNotes 5.0 English)}
        \label{onto-tce}
    \end{table}

    \begin{table}[t]
        \small
        \centering
        \resizebox{\linewidth}{!}{
        \begin{tabular}{lrrr}
            \toprule
            Model & \textsc{All} & \textsc{U-Any} & \textsc{U-Tokens} \\ 
            \midrule
            \textsc{CharLSTM} & 86.38 & 72.71 ($\pm$0.80) & 74.34 ($\pm$0.80) \\ 
            \textsc{CharCNN}  & 86.28 & 72.50 ($\pm$0.76) & 74.10 ($\pm$0.75) \\ 
            \textsc{Flair}    & 89.42 & 77.56 ($\pm$0.21) & 79.05 ($\pm$0.16) \\ 
            \textsc{BERT}     & \textbf{89.84} & \textbf{79.97 ($\pm$0.11)} & \textbf{81.14 ($\pm$0.14)} \\ 
            \textsc{BERT-CRF}    & 89.32 & 78.46 ($\pm$0.56) & 79.63 ($\pm$0.61) \\ 
            \bottomrule
        \end{tabular}}
        \caption{Recall over all mentions and each unseen mention subset (OntoNotes 5.0 English)}
        \label{onto-oov}
    \end{table}

    \paragraph{OntoNotes 5.0 English.}
    Examination of the OntoNotes English data shows that Flair outperforms BERT for type-confusable mentions, but BERT maintains its lead in overall recall when examining unseen mentions.
    Tables~\ref{onto-tce} and \ref{onto-oov} give recall for type-confusable and unseen mentions.\footnote{
    We do not display results for \textsc{TCM-Unseen} and \textsc{Unseen-Type} as they each represent less than 1\% of the test mentions.
    BERT's recall for \textsc{TCM-Unseen} mentions is 19.51 points higher than any other system.
    However, as there are 41 mentions in that set, the difference is only 8 mentions.
    }

    \paragraph{Summary.}
    Table \ref{compare} gives a high-level comparison between BERT and Flair on English data.
    Using the TMR metrics, we find that the models that attain the highest overall recall may not perform the best on tough mentions.
    However, the results vary based on the entity ontology in use.
    In a head-to-head comparison between Flair and BERT on ReCoNLL English, despite Flair having the highest overall and TCM recall, BERT performs better than Flair on \textsc{Unseen-Type}, suggesting that BERT is better at predicting the type for a mention seen only with other types in the training data.
    In contrast, on OntoNotes 5.0 English, BERT attains the highest recall on \textsc{Unseen} mentions, but performs worse than Flair on TCMs.
    The larger and more precise OntoNotes ontology results in the unseen and type-confusable mentions being different than in the smaller CoNLL ontology. 
    In general, Flair performs consistently better on TCMs while BERT performs better on \textsc{Unseen} mentions.

  \begin{table*}
        \centering
        \resizebox{0.9\linewidth}{!}{
        \begin{tabular}{ll|ccccccc}
        \toprule
             Dataset & Model & \textsc{All} & \textsc{U-Any} & \textsc{U-Tok.} & \textsc{U-Type}& \textsc{TCM-All}& \textsc{TCM-Seen} & \textsc{TCM-Unseen}\\
              \hline
            ReCoNLL-English & \textsc{BERT} &  & &  & \ding{52} & &\\
            & \textsc{Flair} & \ding{52} & \ding{52} & \ding{52} &  & \ding{52} & \ding{52} & \ding{52}\\
            \hline
          Ontonotes 5.0 & \textsc{BERT} & \ding{52} & \ding{52} & \ding{52} & N/A & & & N/A\\
            & \textsc{Flair} & &  & & N/A & \ding{52} & \ding{52} & N/A\\
            \bottomrule
        \end{tabular}}
        \caption{Performance comparison between BERT and Flair on English data. A \ding{52} indicates higher recall under a metric. No comparisons are made for \textsc{Unseen-Type} and \textsc{TCM-Unseen} using OntoNotes due to data sparsity.}
        \label{compare}
    \end{table*}

    \subsection{TMR for CoNLL-02 Spanish/Dutch}

       \begin{table*}[tb!]
        \small
        \centering
        \begin{tabular}{lrrrrr}
            \toprule
            {}    & \textsc{All} & \textsc{U-}     & \textsc{U-} & \textsc{U-} & \textsc{TCM-} \\
            Model &              & \textsc{Any} & \textsc{Tok.} & \textsc{Type} & \textsc{All} \\
            \midrule
            \textsc{CharLSTM} & 79.76  & 70.56 ($\pm$0.93)  & 71.72 ($\pm$0.94) & 46.25 ($\pm$3.86) & 70.31 ($\pm$0.84) \\
            \textsc{CharCNN}  & 77.05 & 67.28 ($\pm$0.69) & 68.13 ($\pm$0.51) & 49.38 ($\pm$4.76) & 68.48 ($\pm$0.68)\\
            \textsc{Flair}    & 87.47  & 79.89 ($\pm$0.59) & 81.65 ($\pm$0.50) & 42.81 ($\pm$3.05) & 77.02 ($\pm$1.23)\\
            \textsc{BERT}     & \textbf{88.85} & \textbf{83.04 ($\pm$0.58)} & \textbf{84.55 ($\pm$0.58)} & \textbf{51.25 ($\pm$3.39)} & \textbf{80.00 ($\pm$0.78)}\\
            \textsc{BERT-CRF}    & 88.70 & 82.36 ($\pm$0.42) & 83.93 ($\pm$0.40) & 49.38 ($\pm$1.78) & 79.74 ($\pm$0.63)\\
            \bottomrule
        \end{tabular}
        \caption{Recall over all mentions and unseen and type-confusable mention subsets (CoNLL-2002 Spanish)}
        \label{spanish-oov-tce}
    \end{table*}

       \begin{table}[tb!]
        \small
        \centering
        \resizebox{\linewidth}{!}{
        \begin{tabular}{lrrrr}
            \toprule
            Model & \textsc{All} & \textsc{U-Any} & \textsc{U-Tokens}\\
            \midrule
            \textsc{CharLSTM} & 77.35& 66.32 ($\pm$0.23) & 66.46 ($\pm$0.23)\\
            \textsc{CharCNN}  & 74.55 & 64.50 ($\pm$0.37) & 64.61 ($\pm$0.32)\\
            \textsc{Flair}    & 89.43 & 82.86 ($\pm$0.26) & 83.00 ($\pm$0.26)\\
            \textsc{BERT}     & \textbf{91.68} & \textbf{86.65 ($\pm$0.17)} & \textbf{86.74 ($\pm$0.20)}\\
            \textsc{BERT-CRF}    & 91.26 & 85.88 ($\pm$0.58) & 85.94 ($\pm$0.58)\\
            
            \bottomrule
        \end{tabular}
        }
        \caption{Recall over all mentions and unseen mention subsets (CoNLL-2002 Dutch)}
        \label{dutch-oov-tce}
    \end{table}
    
    Tables~\ref{spanish-oov-tce} and \ref{dutch-oov-tce} give recall for type-confusable and unseen mentions for CoNLL-2002 Spanish and Dutch.\footnote{
      In Table~\ref{spanish-oov-tce}, \textsc{TCM-Unseen} is not shown because it includes less than 1\% of the test mentions (0.6\%); in Table~\ref{dutch-oov-tce} \textsc{Unseen-Type} (0.2\%) and \textsc{TCM} (0.2\%) are not shown.
    }
    The range of the overall recall for Spanish (11.80) and Dutch (17.13) among the five systems we evaluate is much larger than in English (1.76), likely due to systems being less optimized for those languages.
    In both Spanish and Dutch, BERT has the highest recall overall and in every subset.

    While our proposed TMR metrics do not help differentiate models in Spanish and Dutch, they can provide estimates of performance on subsets of tough mentions from different languages and identify areas for improvement.
    For example, while the percentage of \textsc{Unseen-Type} mentions in Spanish (1.8) and ReCoNLL English (1.5) is similar, the performance for BERT for those mentions in Spanish is 34.04 points below that for ReCoNLL English.
    By using the TMR metrics, we have identified a gap that is not visible by just examining overall recall.

    Compared with ReCoNLL English (6.3\%) and Spanish (10.7\%), there are far fewer type-confusable mentions in Dutch (0.2\%).
    Given the sports-centric nature of the English and Spanish datasets, which creates many LOC/ORG confusable mentions, it is likely that their TCM rate is artificially high.
    However the near-zero rate in Dutch is a reminder that either linguistic or data collection properties may result in a high or negligible number of TCMs.
    OntoNotes English shows a similar rate (7.7\%) to ReCoNLL English, but due to its richer ontology and larger set of types, these numbers are not directly comparable.

\section{Conclusion}
    We have proposed Tough Mentions Recall (TMR), a set of evaluation metrics that provide a fine-grained analysis of different sets of formalized mentions that are most challenging for a NER system.
    By looking at recall on specific kinds of ``tough'' mentions---unseen and type-confusable ones---we are able to better differentiate between otherwise similar-performing systems, compare systems using dimensions beyond the overall score, and evaluate how systems are doing on the most difficult subparts of the NER task.

    We summarize our findings as follows.
    For English, the TMR metrics provide greater differentiation across systems than overall recall and are able to identify differences in performance between BERT and Flair, the best-performing systems in our evaluation.
    Flair performs better on type-confusable mentions regardless of ontology, while performance on unseen mentions largely follows the overall recall, which is higher for Flair on ReCoNLL and for BERT on OntoNotes.

    In Spanish and Dutch, the TMR metrics are not needed to differentiate systems overall, but they provide some insight into performance gaps between Spanish and English related to \textsc{Unseen-Type} mentions.

    One challenge in applying these metrics is simply that there may be relatively few unseen mentions or TCMs, especially in the case of lower-resourced languages.
    While we are interested in finer-grained metrics for lower-resourced settings, data sparsity issues pose great challenges.
    As shown in Section~\ref{sec:data}, even in a higher-resourced setting, some subsets of tough mentions include less than 1\% of the total mentions in the test set.
    We believe that lower-resourced NER settings can still benefit from our work by gaining information on pretraining or tuning models towards better performance on unseen and type-confusable mentions.

    For new corpora, these metrics can be used to guide construction and corpus splitting to make test sets as difficult as possible, making them better benchmarks for progress.
    We hope that this form of scoring will see wide adoption and help provide a more nuanced view of NER performance.

\section*{Acknowledgments}

Thanks to two anonymous EACL SRW mentors, three anonymous reviewers, and Chester Palen-Michel for providing feedback on this paper.

\bibliography{anthology,eacl2021}
\bibliographystyle{acl_natbib}

\appendix

    \section{Additional Tables}
    \label{appendix-all}
    
    Please see the following pages for additional tables.

    \begin{table*}[tbh!]
        \centering
        \small
            \begin{tabular}{lrrr}
                \toprule
                {} & Precision & Recall & F1 \\
                \midrule
                \textsc{CharCNN}  & 76.74 ($\pm$0.36) & 74.55 ($\pm$0.27) & 75.63 ($\pm$0.26)\\
                \textsc{CharLSTM} & 78.21 ($\pm$0.34) & 77.35 ($\pm$0.21) & 77.78 ($\pm$0.27) \\
                \textsc{Flair}    & 90.11 ($\pm$0.15) & 89.43 ($\pm$0.13) & 89.77 ($\pm$0.14) \\
                \textsc{BERT}     & \textbf{91.26 ($\pm$0.23)} & \textbf{91.68 ($\pm$0.18)} & \textbf{91.47 ($\pm$0.18)} \\
                \textsc{BERT-CRF}    & 90.75 ($\pm$0.47) & 91.26 ($\pm$0.18) & 91.00 ($\pm$0.32) \\
                \bottomrule
            \end{tabular}
        \caption{Standard precision/recall/F1 for all types for each
        model trained on the CoNLL-2002 Dutch dataset}
        \label{standard_all_type_dutch}
    \end{table*}

    \begin{table*}[tbh!]
        \centering
        \small
            \begin{tabular}{lrrr}
                \toprule
                {} & Precision & Recall & F1 \\
                \midrule
                \textsc{CharCNN}  & 77.75 ($\pm$0.22) & 77.05 ($\pm$0.21) & 77.40 ($\pm$0.20)\\
                \textsc{CharLSTM} & 80.09 ($\pm$0.59) & 79.76 ($\pm$0.63) & 79.93 ($\pm$0.61) \\
                \textsc{Flair}  & 86.96 ($\pm$0.23) & 87.47 ($\pm$0.19) & 87.21 ($\pm$0.20)\\
                \textsc{BERT}   & \textbf{87.36 ($\pm$0.52)} & \textbf{88.85 ($\pm$0.39)} & \textbf{88.10 ($\pm$0.45)} \\
                \textsc{BERT-CRF}   & 87.25 ($\pm$0.38) & 88.70 ($\pm$0.20) & 87.97 ($\pm$0.29)\\
                \bottomrule
            \end{tabular}
        \caption{Standard precision/recall/F1 for all types for each
        model trained on the CoNLL-2002 Spanish dataset}
        \label{standard_all_type_spanish}
    \end{table*}

    \begin{table*}[tbh!]
        \centering
        \small
        \begin{tabular}{lrrrrrrrrr}
            \toprule
            {Mentions} & ALL & GPE & PER & ORG & DATE & CARD & NORP & PERC & MONEY \\
            \midrule
     
            \textsc{Unseen-Any}      & 30.3 & 10.5 & 48.9 & 41.4 & 20.3 & 15.3 & 12.4 & 29.5 & 61.8 \\
            \textsc{Unseen-Tokens} & 29.4 & 9.9 & 48.0 & 40.8 & 19.7 & 14.9 & 12.0 & 29.5 & 60.2 \\
            \textsc{Unseen-Type}     & 0.9 & 0.6 & 0.9 & 0.6 & 0.6 & 0.4 & 0.4 & 0.0 & 1.6 \\
            \midrule
            \textsc{TCM-All}    & 7.7 & 11.5 & 1.7 & 4.9 & 3.2 & 15.4 & 18.5 & 0.0 & 5.1 \\
            \textsc{TCM-Seen}   & 7.3 & 11.1 & 1.6 & 3.8 & 3.2 & 15.2 & 18.4 & 0.0 & 5.1 \\
            \textsc{TCM-Unseen} & 0.4 & 0.4 & 0.1 & 1.1 & 0.1 & 0.2 & 0.1 & 0.0 & 0.0 \\
            \midrule
            Total (Count)   & 11,265 & 2,241 & 1,991 & 1,795 & 1,604 & 936 & 842 & 349 & 314 \\

            \bottomrule
        \end{tabular}
        \caption{Percentage of all mentions in each subset, with total mentions in the final row (OntoNotes 5.0 English). Due to space constraints, types are split across this table and the following one.}
        \label{onto-percentage1}
    \end{table*}

    \begin{table*}[tbh!]
        \centering
        \small
        \begin{tabular}{lrrrrrrrrrr}
            \toprule
            {Mentions} & TIME & ORD & LOC & WA & FAC & QUAN & PROD & EVENT & LAW & LANG \\
         
            \midrule
            \textsc{Unseen-Any}      & 41.5 & 3.6 & 39.1 & \textbf{83.1} & 80 & 73.3 & 52.6 & 47.6 & 75.0 & 22.7 \\
            \textsc{Unseen-Tokens} & 39.6 & 3.1 & 34.1 & \textbf{78.9} & 74.8 & 73.3 & 48.7 & 47.6 & 57.5 & 4.5 \\
            \textsc{Unseen-Type}     & 1.9 & 0.5 & 5.0 & 4.2 & 5.2 & 0.0 & 3.9 & 0.0 & \textbf{17.5} & 18.2 \\
            \midrule
            \textsc{TCM-All}    & 7.5 & 12.8 & 14 & 5.4 & 15.5 & 0.0 & 0.0 & 7.9 & 0.0 & \textbf{54.5} \\
            \textsc{TCM-Seen}   & 7.5 & 12.8 & 14 & 2.4 & 14.8 & 0.0 & 0.0 & 7.9 & 0.0 & \textbf{54.5} \\
            \textsc{TCM-Unseen} & 0.0 & 0.0 & 0.0 & \textbf{3.0} & 0.7 & 0.0 & 0.0 & 0.0 & 0.0 & 0.0 \\
               \midrule
            Total (Count)   & 212 & 195 & 179 & 166 & 135 & 105 & 76 & 63 & 40 & 22 \\
            \bottomrule
        \end{tabular}
        \caption{Percentage of all mentions in each subset, with total mentions in the final row (OntoNotes 5.0 English). Due to space constraints, types are split across this table and the preceding one.}
        \label{onto-percentage2}
    \end{table*}

\end{document}